\documentclass[conference]{IEEEtran}
\IEEEoverridecommandlockouts
\usepackage{cite}
\usepackage{amsmath,amssymb,amsfonts}
\usepackage{algorithmic}
\usepackage{graphicx}
\usepackage{textcomp}
\usepackage{xcolor}
\def\BibTeX{{\rm B\kern-.05em{\sc i\kern-.025em b}\kern-.08em
    T\kern-.1667em\lower.7ex\hbox{E}\kern-.125emX}}

\usepackage{epsfig}
\usepackage{graphicx}
\usepackage{amsmath}
\usepackage{amssymb}

\usepackage[utf8]{inputenc} % allow utf-8 input
\usepackage[T1]{fontenc}    % use 8-bit T1 fonts
\usepackage{hyperref}       % hyperlinks
\usepackage{booktabs}       % professional-quality tables
\usepackage{amsfonts}       % blackboard math symbols
\usepackage{nicefrac}       % compact symbols for 1/2, etc.
\usepackage{microtype}      % microtypography
\usepackage{xcolor}         % colors
\usepackage{pifont}

\usepackage{graphicx,epstopdf}
\usepackage{amssymb}
\usepackage{amsmath}
\usepackage{amsthm}

\usepackage{amsthm,amsmath,amssymb}
\usepackage{mathrsfs}

\usepackage{tabu}
\usepackage{multirow}
\usepackage{multicol}
\usepackage{makecell}
\usepackage{booktabs}

\newtheorem{Theorem}{Theorem}
\newtheorem{myDef}[Theorem]{Definition}
\newtheorem{myRemark}{\textbf{Remark}}

\usepackage[ruled]{algorithm2e} % For algorithms

\begin{document}

\title{Parameter Hierarchical Optimization for Visible-Infrared Person Re-Identification\\
}

\author{\IEEEauthorblockN{Zeng Yu}
\IEEEauthorblockA{\textit{School of Computer Science and
Artificial Intelligence} \\
\textit{Southwest Jiaotong University}\\
Chengdu, China\\
zyu@swjtu.edu.cn}
\and
\IEEEauthorblockN{Yunxiao Shi}
\IEEEauthorblockA{\textit{School of Computer Science and
Artificial Intelligence} \\
\textit{Southwest Jiaotong University}\\
Chengdu, China\\
headre@my.swjtu.edu.cn}
}

\maketitle

\begin{abstract}

Visible-infrared person re-identification (VI-reID) aims at matching cross-modality pedestrian images captured by disjoint visible or infrared cameras. Existing methods alleviate the cross-modality discrepancies via designing different kinds of network architectures. Different from available methods, in this paper, we propose a novel parameter optimizing paradigm, parameter hierarchical optimization (PHO) method, for the task of VI-ReID. It allows part of parameters to be directly optimized without any training, which narrows the search space of parameters and makes the whole network more easier to be trained. Specifically, we first divide the parameters into different types, and then introduce a self-adaptive alignment strategy (SAS) to automatically align the visible and infrared images through transformation. Considering that features in different dimension have varying importance, we develop an auto-weighted alignment learning (AAL) module that can automatically weight features according to their importance. Importantly, in the alignment process of SAS and AAL, all the parameters are immediately optimized with optimization principles rather than training the whole network, which yields a better parameter training manner. Furthermore, we establish the cross-modality consistent learning (CCL) loss to extract discriminative person representations with translation consistency.
We provide both theoretical justification and empirical evidence that our proposed PHO method outperform existing VI-reID approaches.

%Experimental results on image- and video-based VI-ReID datasets demonstrate the effectiveness of our approach.
\end{abstract}

\section{Introduction}

Person re-identification (reID) aims at matching pedestrian images captured from multi-disjoint cameras \cite{ye2021deep,yu2018unsupervised,zheng2016person}.
It has developed into an important branch of image retrieval in recent years due to its wide range of successful applications, including  intelligent surveillance, public security, and pedestrian analysis \cite{liao2021transmatcher,tay2019aanet,xu2018attention}. Most reID techniques focus on single-modality retrieval tasks \cite{eom2019learning,qi2023unsupervised,wang2022nformer,zheng2019joint}, e.g., all the pedestrian images are acquired by visible cameras in the daytime.  Nevertheless, visible cameras are unable to obtain clear images in dark conditions. To overcome this obstacle, infrared cameras are adopted in many surveillance scenarios, which are robust to poor-illumination environment and provide effective complement to visible cameras. Here,
visible images have three channels containing rich color information, while infrared images have one channel capturing information of invisible light.  Therefore, in practice, we usually need to match visible and infrared images, raising the task of visible-infrared person re-identification (VI-reID) \cite{sun2023counterfactual,wu2017rgb}.

VI-reID suffers from two major challenges, the large cross-modality discrepancies caused by different imaging mechanism of visible and infrared images, and the intra-class variations arisen from human pose changing, different viewpoint, background occluding, etc. Existing VI-reID approaches mainly focus on decreasing the cross-modality discrepancies utilizing metric learning \cite{feng2019learning,fu2021cm,ye2018visible,ye2018hierarchical} along with robust feature representations \cite{dai2018cross,si2023tri}, and further improve the performance with memory-based augmentation \cite{li2022visible,liu2022learning} or disentanglement techniques \cite{choi2020hi,pu2020dual}. These methods based on convolutional neural networks (CNNs) learn coarse image-level or part-level person representations and assume that visible and infrared images are roughly aligned. However, recent studies demonstrate that person representations between visible and infrared images are not well-aligned \cite{park2021learning,wang2019rgb,wan2023g2da}. Actually, misaligned features have an adverse effect on alleviating the cross-modality discrepancies and reduce the accuracy of person representations learning.

\begin{figure*}[t]
%\fbox{\rule[-.5cm]{0cm}{4cm} \rule[-.5cm]{4cm}{0cm}}
\begin{center}
%\centerline{\includegraphics[width=14cm]{Drnet.pdf}}
\centerline{\includegraphics[width=14cm]{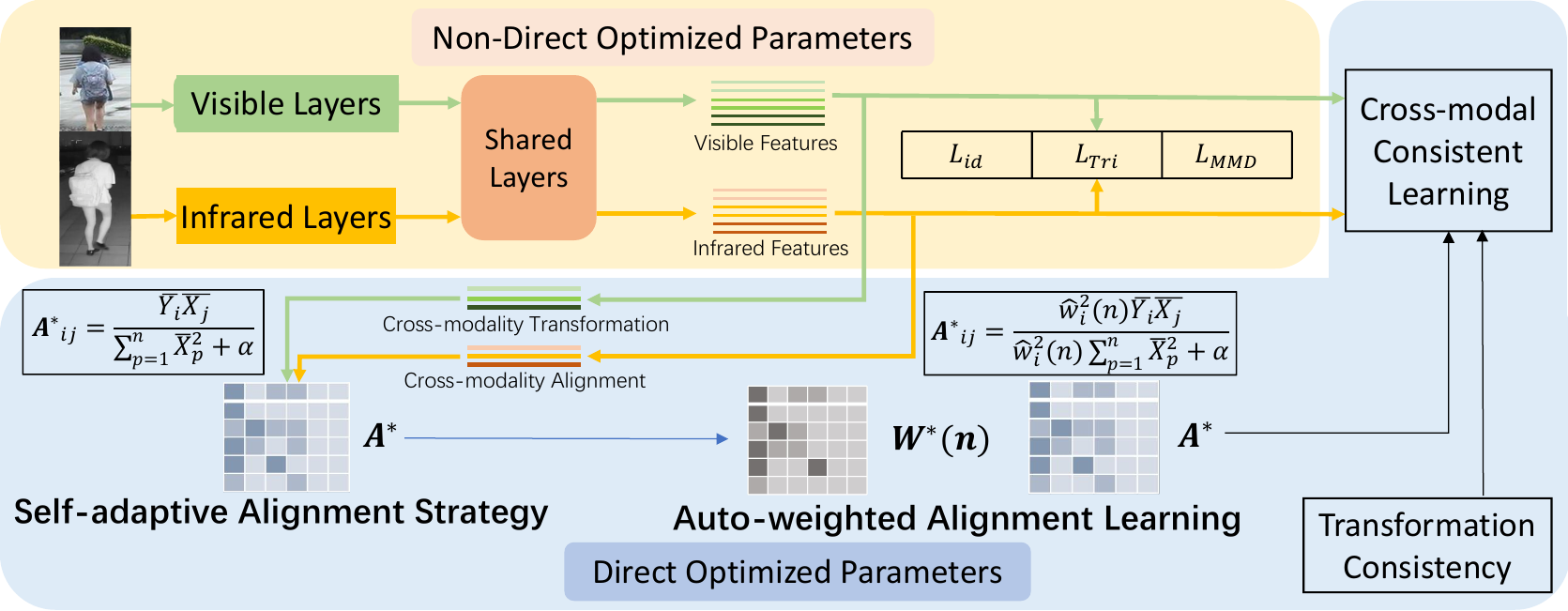}}
\caption{Framework of our PHO method. All the parameters are first divided into non- and direct optimized parameters. Then the direct optimized parameters are immediately optimized with self-adaptive alignment strategy and auto-weighted alignment learning, and the remaining non-direct optimized parameters need to be trained with optimizers. Because partial parameters are directly optimized instead of training, it can reduce the search space of parameters and result in easier training.}
\label{flame}
\end{center}
\end{figure*}

In this paper, we propose a parameter hierarchical optimization (PHO) approach to address the issue of cross-modality discrepancies. More specifically, all the parameters in our network are first divided into two types: one needs to be trained with optimizers, such as stochastic gradient descent (SGD) or Adam, and the other can be directly optimized without any training (see Definition \ref{def1} for details). Then a self-adaptive alignment strategy (SAS), which can automatically align the visible and infrared images through translating visible/infrared images into infrared/visible images, is introduced for eliminating the cross-modality discrepancies. It should be pointed out that parameters in translation stage are independently optimized with optimization rules, but not training the whole network. Considering that features in different dimension have varying importance, an auto-weighted alignment learning module is developed. In order to guarantee same persons close to each other rather than different persons after translation, we also introduce cross-modality consistent learning in the paper. The overall framework of our algorithm can be seen in Fig. \ref{flame}.
%首先将网络中的参数分为可以学习的（训练的）和不需要训练的（直接优化的）参数
%提出了一种新的训练参数模式
%减小模态差异，和配准都解决了
%需要强调：这里的先优化，通过最优化准则，给出部分参数的最优近似值，然后，再通过网络学习来得到其他最优值
%注：先对某些可以优化的参数进行独立求解，让这些参数固定下来，这样减少了参数的搜索空间，使得参数空间更容易被优化。
%不仅在图像还在视频上也取得好的结果

The main contributions of this paper can be summarized as follows:

\begin{itemize}
\item We propose the PHO method to train the whole network, which presents a new manner to optimize the parameters. In PHO, partial parameters can be obtained with optimization principles instead of training, which reduces the search space of parameters and makes it more easier to optimize when training. To our best knowledge, this is the first work that trains the neural network with a hierarchical parameter way for reID.
\item We introduce the SAS to automatically align the visible and infrared images, which also alleviates the discrepancies between multi-modality images effectively. In order to make the alignment more accurate, an auto-weighting module is proposed, which can automatically weight features according to the importance.
\item We establish the cross-modality consistent learning loss to ensure sufficient extracting the discriminative person representations by using cross-modal translation consistency.
\item Our method is evaluated on both image- and video-based VI-reID datasets. Experimental results show that the effectiveness and efficiency of our approach.

\end{itemize}

\section{Related Work}
\label{gen_inst}

\paragraph{VI-reID.} VI-reID attempts to match pedestrian images between different modalities, which has received remarkable progress in recent years \cite{chen2022structure,hao2019hsme,ye2020visible,ye2018hierarchical,zhang2021attend}. Wu et al. \cite{wu2017rgb} collected the first cross-modality dataset named SYSU-MM01 and proposed a zero-padding approach.
%Ye et al. \cite{ye2018visible} designed a dual-path network and introduced a bi-directional dual-constrained top-ranking loss to learn robust person representations for VI-reID.
To obtain discriminative feature representations, a lot of metric learning techniques have been developed \cite{lu2020cross,ye2019bi,ye2020visible,wang2019learning}, which is robust to cross-modality dataset and achieves significantly better performance. To use the traditional triplet loss, recent works designed a cross-modal triplet loss, where the anchor and positive/negative pairs are sampled from different modalities \cite{feng2019learning,wang2020deep,wei2021flexible}. In addition, a variety of methods based on generative adversarial networks (GAN) are adopted to remit pixel-level modality discrepancy \cite{choi2020hi,wang2019rgb,wang2020cross,zhang2021rgb}. For example, Wang et al. \cite{wang2019rgb} used GAN to translate pedestrian images from one modality to another and bridged the cross-modality gap via joint pixel and feature alignment. Other GAN-based approaches also utilized instance-level alignments \cite{wang2020cross}, feature disentanglement \cite{choi2020hi}, or generated intermediate modality \cite{li2020infrared} to further narrow the discrepancies of different modalities.
Considering that video sequences contain much richer spatial and temporal information, Lin et al. \cite{lin2022learning} constructed a video-based VI-reID dataset. Despite the great success of VI-reID methods, we find that a large number of parameters need to be trained and cross-modality alignment is also very challenging. Hence, in the paper, we focus on providing some insights to reduce the number of parameters for training and improve the performance of alignment-based approach.

\paragraph{Alignment-based VI-reID.} The alignment-based VI-reID approaches mainly exist three lines of literature: (\textrm{i}) pixel alignment-based methods \cite{choi2020hi,wu2017rgb,wang2019learning}; (\textrm{ii}) feature alignment-based methods \cite{hao2019hsme,ye2018hierarchical,ye2020dynamic,zhang2021attend}; (\textrm{iii}) sample alignment-based methods \cite{wang2020cross,wang2020deep}. Pixel alignment-based approaches mitigate modality discrepancy from pixel level, often use GAN to translate input images into their corresponding modality. However, adversarial training is unstable and requires lots of parameters for training \cite{kansal2020sdl,ye2020visible}. Feature alignment-based methods try to alleviate such heterogeneity with cross-modality feature alignment.
Normally, they first leverage modality-specific CNNs networks to map visible and infrared images into a common feature space, and then design a variety of constraints, such as triplet loss or consistent learning loss, to restrict the whole network to obtain discriminative representations. Recent studies employ probability statistics methods such as maximum mean discrepancy (MMD) \cite{jambigi2021mmd,zheng2022progressive} for modality feature alignment and achieve remarkable performance. Sample alignment-based methods eliminate modality divergence within class via aligning different modality images from sample level. Compared with current alignment-based VI-reID methods, we also address the cross-modality discrepancies in a feature level. To this end, a self-adaptive alignment strategy is introduced to automatically align the visible and infrared images explicitly, which also allows discriminative feature learning even for misaligned images.

\paragraph{Parameter Optimization.} Parameter optimization is very important for a deep learning based method, its quality is directly related to the success or failure of a new algorithm. However, existing VI-reID methods train the parameters of whole neural network either with SGD or Adam \cite{kingma2014adam,reddi2019convergence}. There are no related works involved in decreasing the number of training parameters. In the paper, we propose the PHO method, in which part of the parameters are independently optimized without training, to narrow the parameters space and make it easier to search for the optimal values. Our implementation and idea of training parameters are very different from available methods, which may provide a new perspective to further improve these studies.

%定义：参数层级，一种是可以直接优化的，一种是通过学习给出的

\section{Proposed Method}
\label{Hier}
%为了解决模态差异大和不匹配问题
%In this section, we provide the details of the proposed PHO

\subsection{Hierarchical Parameters}

In order to effectively train the whole network, we first divide the parameters of neural network into different types. Then we obtain the optimal value of partial parameters through solving the optimal constraints without training, which makes the remaining training process more effective. The hierarchical parameters and different types of parameters are defined as follows.

\begin{myDef} \label{def1}
(Hierarchical Parameters).
In a neural network, if the optimal value of parameters can be directly calculated with optimization principles instead of training, we call them direct optimized parameters; If the optimal value of parameters can be obtained through training, we call them learnable parameters;
If the optimal value of parameters can not be directly optimized and need to be trained with optimizers, we call them non-direct optimized parameters. The parameters in a neural network are hierarchical parameters if all the parameters are composed of non- and direct optimized parameters.
\end{myDef}

Obviously, direct optimized parameters are also learnable parameters. As shown in experiments, we can also acquire the optimal value of direct optimized parameters via training the whole network, which gets poorer performance than that of directly optimizing parameters. The main reason is that directly optimizing parameters can narrow the search space of parameters and make the whole network easier to be trained.

\begin{myRemark}
If a neural network has hierarchical parameters, the search space of parameters can be reduced and the training task may become easier.
\end{myRemark}

This prompts us to discover the hierarchical parameters of a neural network and yields a better parameter optimization method. It should be pointed out that it is the first work that optimizes the whole network through a parameter hierarchical optimization approach, which provides a new idea to further improve other methods.

\subsection{Self-adaptive Alignment Strategy (SAS)}

To alleviate the cross-modality discrepancies, we propose a self-adaptive alignment strategy to automatically match the visible and infrared images explicitly. Specifically, given a visible image $X_{i}\in \Re^{n}$ and an infrared image $Y_{j}\in\Re^{n}$, we address the cross-modality alignment through translating one modal into another with a mapping $\psi(\cdot)$ and allowing them close to each other. For simplicity of computation, we use a transformation matrix $A\in \Re^{n\times n}$ to approximately replace the mapping $\psi(\cdot)$ and regularize the matrix $A$. Therefore, the issue of cross-modality alignment can be converted to minimize the objective function $P_{1}$ with unknown mapping $A$ as follows:

\begin{equation}
\mathop{\min}\limits_{A} P_{1}(A)= \left\Vert \frac{1}{M} \sum_{i=1}^{M} AX_{i}- \frac{1}{N}\sum_{j=1}^{N} Y_{j} \right\Vert_2^2 + \alpha \left\Vert A\right\Vert_F^2,
\end{equation}

where $\alpha$ is a hyper parameter that controls the strength of the regularization, $M$ and $N$ are the numbers of visible and infrared images in one mini-batch, respectively.

Let $ \bar{X}= \frac{1}{M} \sum_{i=1}^{M} X_{i}$ and $\bar{Y} =\frac{1}{N}\sum_{j=1}^{N} Y_{j}$ be the mean values of visible and infrared images in one mini-batch, respectively. The solution of optimization problem $P_{1}$ is given in Theorem \ref{Theorem1}.

\begin{Theorem} \label{Theorem1}
Problem $P_{1}(A)$ is minimized iff
\begin{equation}
A_{ij}^\star= \frac{ \bar{Y}_{i}\bar{X}_{j}}{  \sum_{p=1}^{n} \bar{X}_{p}^2 + \alpha } \ \ for \ 1\leq i,j\leq n.   \label{solve1}
\end{equation}
where $A^\star= (A_{ij}^\star )_{n\times n}$ be the optimum transformation matrix with respect to problem $P_{1}$, $\bar{X}_{j}$ is the $j$-th dimension of $\bar{X}$, and $\bar{Y}_{i}$ is the $i$-th dimension of $\bar{Y}$.
\end{Theorem}

The proof of this theorem is given in Appendix A. Different from traditional methods, which obtain the transformation matrix through training the whole network, Theorem \ref{Theorem1} presents an new manner to acquire the optimal values of transformation matrix.

\begin{myRemark}
In order to obtain the transformation matrix, Theorem \ref{Theorem1} just solve an optimization problem rather than training the whole network, which yields less parameters for training and thus leads to much easier optimization for the neural network.
\end{myRemark}

\subsection{Auto-weighted Alignment Learning (AAL)}

Considering that features in different dimension have different importance, we propose the auto-weighted alignment learning method for cross-modality feature alignment. It can automatically distinguish the importance of different feature dimension by employing a weighting function which emphasizes important features in alignment process. Concretely, similar to SAS, we tackle the issue of cross-modality alignment
through translating with a transformation matrix $A\in \Re^{n\times n}$ too. Hence, it is also equivalent to minimize the objective function $P_{2}$ with unknown weight matrix $W(n)$ and mapping $A$.

\begin{equation}
\begin{split}
\mathop{\min}\limits_{W(n),A} P_{2}\big(W(n),A\big) & = \left\Vert W(n) \left( \frac{1}{M} \sum_{i=1}^{M} AX_{i}- \frac{1}{N}\sum_{j=1}^{N} Y_{j} \right) \right\Vert_2^2  \\
& + \alpha \left\Vert A\right\Vert_F^2,
\end{split}
\end{equation}

subject to

\begin{equation}
%\begin{displaymath}
\sum_{i=1}^{n}w_{i}(n)=1,\ 0\leq w_{i}(n) \leq 1,
\end{equation}
%对角矩阵
%where $W(n)=diag\big(w_{1}^{\beta}(n),w_{2}^{\beta}(n),...,w_{n}^{\beta}(n)\big)$  %有beta参数的$\beta$ is a smoothing factor for each element of weight matrix $W(n)$

where $W(n)=diag\big(w_{1}(n),w_{2}(n),...,w_{n}(n)\big)$ is a diagonal weight matrix, $\alpha$, $M$ and $N$ are same as problem $P_{1}$.

%In practice, because $n$ is a large number, the value of each $w_{i}(n),i=1,2,...,n$ is a small number. Hence, for convenience of calculations, we change the subject conditions as follows:
%\begin{equation}
%\sum_{i=1}^{n}w_{i}(n)=n,\ 0\leq w_{i}(n),
%\end{equation}
%这里不做理论说明，在后面实验部分加上说明：为了计算方便，适当放大了w_{i}(n)比例

Inspired by \cite{huang2005automated} and \cite{yu2016iterative}, the above optimization problem $P_{2}$ can be tackled by iteratively solving the following two minimization problems:
\begin{itemize}
\item[\quad 1.] Problem $P_{21}$: Fix $W(n)=\hat{W}(n)$ and solve the reduced problem $P_{2}(\hat{W}(n),A)$;
\end{itemize}

\begin{itemize}
\item[\quad 2.] Problem $P_{22}$: Fix $A=\hat{A}$ and solve the reduced problem $P_{2}(W(n),\hat{A})$.
\end{itemize}

In fact, for given $\hat{W}(n)$, problem $P_{21}$ is similar to $P_{1}$. The solution of $P_{21}$ is given in Theorem \ref{Theorem2}.

\begin{Theorem} \label{Theorem2}
Let $W(n)=\hat{W}(n)$ be fixed. $P_{2}(\hat{W}(n),A)$ is minimized iff
\begin{equation}
A_{ij}^\star= \frac{\hat{w}_{i}^{2}(n) \bar{Y}_{i}\bar{X}_{j}}{  \hat{w}_{i}^{2}(n) \sum_{p=1}^{n} \bar{X}_{p}^2 + \alpha  } \ \ for \ 1\leq i,j\leq n.
\end{equation}
where $A^\star= (A_{ij}^\star )_{n\times n}$ be also the optimum transformation matrix but with respect to problem $P_{2}$.
\end{Theorem}

The solution of remaining problem $P_{22}$ is given in following Theorem \ref{Theorem3}.

\begin{Theorem} \label{Theorem3}
Let $A =\hat{A}$ be fixed. $P_{2}(W(n),\hat{A})$ is minimized iff
\begin{equation}
  w_{i}^\star(n)= \left\{\begin{array}{cc}
  0 & \textrm{if $E_{i}=0$}\\
  $$ \frac{1}{\mathop{ \sum_{t=1}^{n}}\limits_{E_{t}\neq 0} \frac{E_{i}}{E_{t}}}  & \textrm{if $E_{i}\neq 0$} $$    \label{weight}
  \end{array} \right.
\end{equation}
  where
  \begin{equation}
  E_{i}=\left(\sum_{j=1}^{n} \hat{A}_{ij}  \bar{X}_{j} - \bar{Y}_{i} \right)^{2}    \label{error}
  \end{equation}
\end{Theorem}

The proofs of Theorem \ref{Theorem2} and \ref{Theorem3} are provided in Appendix. Given the visible and infrared features, the principal for feature weighting is to assign a larger weight to a dimension of feature that has a smaller alignment error
and a smaller one to a dimension of feature that has a larger alignment error. The alignment error is given by (\ref{error}) and the weight $w_{i}^\star(n)$ is calculated by (\ref{weight}) in Theorem \ref{Theorem3}. The algorithm of auto-weighted alignment learning can be found in the supplementary material.

\begin{myRemark}
Theorem \ref{Theorem2} and \ref{Theorem3} demonstrate that the parameters of transformation matrix and weight matrix can also be solved with optimization principles rather than training the whole network, which yields a better manner of parameters training.
\end{myRemark}

\subsection{Cross-modality Consistent Learning Loss (CCL)}

We have shown that the alignment and discrepancies of cross-modality can be handled with SAS and AAL by employing an optimum transformation matrix $A^\star$ and an auto-weighted matrix $W^\star{(n)}$. However, after transformation, we can not guarantee the network to obtain an effective person representation. To learn discriminative feature representations of cross-modality, we introduce the cross-modality consistent learning loss.
Let $\bar{X}_{c}=\frac{1}{N_{p}}\sum_{p=1}^{N_{p}} X_{cp}$ and $\bar{Y}_{c}=\frac{1}{N_{q}}\sum_{q=1}^{N_{q}}Y_{cq}$ be the mean values,
$N_{p}$ and $N_{q}$ be the number of visible and infrared data belonging to the $c$-th class, respectively. With the optimum transformation matrix $A^\star$, the intra-class variance is formulated as:

\begin{equation}
\begin{split}
\mathcal{L}_{intra} & =
\frac{1}{N_{c}} \sum_{c=1}^{N_{c}} \left[ \| A^\star  \bar{X}_{c} - \bar{Y}_{c}  \|_{2}  \right. \\
& \left. +\beta  \left(  \frac{1}{N_{p}} \sum_{p=1}^{N_{p}} \| A^\star X_{cp}  - \bar{Y}_{c}\|_{2}  + \frac{1}{N_{q}} \sum_{q=1}^{N_{q}} \| A^\star \bar{X}_{c}-Y_{cq} \|_{2}  \right)\right]
\end{split}
\end{equation}

where $\beta$ is a hyper parameter and $N_{c}$ is the number of classes. The intra-class loss encourages the visible and infrared images to close to each other if they acquired from the same person. Inspired by triple loss, we develop the CCL loss, which is also involved in inter-class variance.
%To separate the person with different identity as much as possible, the inter-class variance is given.

\begin{equation}
\mathcal{L}_{CCL} = \left[ \rho + \gamma \mathcal{L}_{intra} - \mathop{\min}\limits_{1\leq i,j \leq N_{c}, i\neq j}  \| A^\star  \bar{X}_{i} - \bar{Y}_{j}   \|_{2} \right]_{+}
\end{equation}

where $\rho$ and $\gamma$ are hyper parameters. After translation, the CCL loss makes the same person with different modalities close to each other rather than different persons.

%\paragraph{Kernelization:} For nonlinear problems, consider kernel mapping and kernel matrix  We utilize the  theorem to formulate

\subsection{Overall Objective Function}
\label{Overall}

To further enhance the discriminative and modality-shared person representations, we employ the typically used two-stream
network with ResNet50 \cite{he2016deep} as the backbone \cite{park2021learning,ye2018visible,wang2020cross}, to tackle the different modalities. Specifically, the shallow CNNs layers separately learn modal-specific features for visible and infrared images/sequences, and the deep layers share the weights to extract modal-invariant features of cross-modality. For image- or video-based VI-ReID dataset, we use different objective functions as the baseline $\mathcal{L}_{base}$. By following \cite{jambigi2021mmd}, the baseline learning loss $\mathcal{L}_{base}$ for image-based VI-ReID can be formulated as:

\begin{equation}
\mathcal{L}_{base} = \mathcal{L}_{id}  + \mathcal{L}_{Hc-Tri} + \mathcal{L}_{MMD}
\end{equation}

where $\mathcal{L}_{id}$, $\mathcal{L}_{Hc-Tri}$ and $\mathcal{L}_{MMD}$ are the identity loss, hetero-center triplet loss and maximum mean discrepancy loss, respectively.

However, for video-based VI-ReID \cite{lin2022learning}, the baseline training loss $\mathcal{L}_{base}$ is described as follows:

\begin{equation}
\mathcal{L}_{base} = \mathcal{L}_{id} + \mathcal{L}_{Tri} +  \mathcal{L}_{adv}
\end{equation}

where $\mathcal{L}_{id}$, $\mathcal{L}_{Tri}$ and $\mathcal{L}_{adv}$ are the identity loss, triplet loss and adversarial loss, respectively.

Therefore, for image- or video-based VI-ReID, the overall objective function of our proposed approach can be formulated as in Eq. (\ref{totalLoss}):

\begin{equation}
\label{totalLoss}
\mathcal{L}_{total} = \mathcal{L}_{base} + \lambda \mathcal{L}_{CCL}
\end{equation}

It should be pointed out that the cross-modality alignment learning of SAS and AAL is not included in the overall training loss. That's because the parameter optimization of cross-modality alignment process is directly optimized with optimization principles instead of training, which  can narrow the search space of parameters and let training task easier to find the optimal parameters. For more details about how to optimize parameters hierarchically, please refer to Algorithm \ref{algAAL}.

\begin{algorithm}[htbp]  %[t]
\SetAlgoNoLine
\KwIn{Training visible images $X$ and infrared images $Y$, learning rate $\eta$, tradeoff coefficients $\alpha$, $\beta$, $\gamma$ and $\lambda$.}
\textbf{1. Separate} all parameters into direct optimized parameters $A$ and $W(n)$, and non-direct optimized parameters $\theta$ according to Definition \ref{def1} \;
\textbf{2. Initialize} the non-direct optimized parameters $\theta$ \;
\Repeat{Stopping criteria are met}{
        \textbf{3. Select} $M$ visible and $N$ infrared samples from $X$ and $Y$, respectively\;
        \textbf{4. Calculate} optimized $A^\star$ and $W^\star{(n)}$ with SAS or AAL; \quad   \, \!  \! \! \!  \# The first objective function  \

        \textbf{5. Define} the overall training loss $\mathcal{L}_{total} = \mathcal{L}_{base} + \lambda \mathcal{L}_{CCL}$;   \, \, \,         \# The second objective function  \

        \textbf{6. Update} parameters $\theta$ by $\theta \longleftarrow \theta - \eta  \frac{\partial \mathcal{L}_{total}}{\partial \theta}$.
      }
\caption{The PHO Algorithm}
\label{algAAL}
\end{algorithm}

\section{Experiments}

%To show the effectiveness of our proposed method, we conduct experiments on two standard benchmarks including SYSU-MM01 \cite{wu2017rgb} and RegDB \cite{nguyen2017person}. We also evaluate our approach on a video-based VI-ReID dataset HITSZ-VCM \cite{lin2022learning}. In addition,
%we perform ablation studies on different parameter optimization strategies and network architectures.

\subsection{Dataset and Evaluation Protocol}

SYSU-MM01 \cite{wu2017rgb} is a pioneer benchmark for the task of VI-ReID. This dataset contains 395 identities for training, which consists of 22,258 visible images captured from four visible cameras and 11,909 near-infrared images taken by two near-infrared cameras. For testing,
there are 96 identities with 3,803 near-infrared images for a query set and 301 randomly sampled visible images for a gallery set.
RegDB \cite{nguyen2017person} contains 412 identities, which acquired from a dual system of visible and thermal cameras. The identities are randomly and equally chosen as the training set with 2,060 images from 206 person identities and the remaining 2,060 images with 206 identities compose the testing set. HITSZ-VCM \cite{lin2022learning} is a unique video-based VI-ReID dataset, which collects 927 person identities with 251,452 visible images captured by 6 visible cameras and 211,807 infrared images acquired from 6 infrared cameras. In HITSZ-VCM, every 24 consecutive frames constitute a tracklet. Following \cite{lin2022learning}, we divide HITSZ-VCM into training set with 11,061 tracklets of 500 identities and testing set with 10,802 tracklets of 427 identities.

To evaluate the proposed method, cumulative matching characteristic curve (CMC) and mean average precision (mAP) are adopted as the evaluation metrics. The CMC curve reports the probability that matches the correct persons in top-k retrieval results, and mAP is a comprehensive evaluation performance of multi-matching.

\subsection{Implementation Details}

Following existing VI-ReID methods \cite{jambigi2021mmd,ye2021deep,lin2022learning}, we employ the commonly used two-stream network with ResNet50 as our backbone, and utilize generalized-mean (GeM) pooling \cite{radenovic2018fine} instead of average or max pooling to fuse the frame-level features. All the inputs images are firstly resized to $288\times144$ and padded with 10. Random horizontal flipping and random erasing augmentation \cite{zhong2020random} are applied for data augmentation.  We use a stochastic gradient descent (SGD) with momentum as 0.9 to train our network. We set the initial learning rate to 0.1 for randomly initialized parameters, and 0.01 for pre-trained parameters such as batch normalization and classifier layers. A warm-up strategy \cite{liu2020parameter} is used to improve the performance.
For SYSU-MM01 and RegDB datasets, we train the model for 80 epochs and decay the learning rate by 0.1 at 40-th epoch and 0.01 at 60-th epoch.  In a mini-batch, we randomly sample 6 identities for each modality and chose 6 images for each identity.
For HITSZ-VCM dataset, we train the network 200 epochs and decrease the learning rate by a factor of 10 at the 35-th and 80-th epochs.
We randomly chose 8 identities for each modality, in which every identity has 1 visible and 1 infrared tacklet, and randomly sampled 12 frames for each tracklet.

\subsection{Comparison with State-of-the-Art Methods}

\paragraph{Results on SYSU-MM01 and RegDB.} To demonstrate the superiority of PHO, we first compare our method with current state-of-the-art VI-ReID methods, including Zero-Pad \cite{wu2017rgb}, HCML \cite{ye2018hierarchical}, AlignGAN \cite{wang2019rgb}, Xmodal  \cite{li2020infrared}, DDAG \cite{ye2020dynamic}, LbA \cite{park2021learning}, NFS \cite{chen2021neural}, MC-AWL \cite{ling2021multi}, MID \cite{huang2022modality}, SPOT \cite{chen2022structure}, DART \cite{yang2022learning}, FMCNet \cite{zhang2022fmcnet} and MRCN \cite{zhang2023mrcn}. Table \ref{sysu} reports the rank-1 accuracy (\%) and mAP (\%) for a single-shot setting on the image-based VI-ReID SYSU-MM01 \cite{wu2017rgb} and RegDB \cite{nguyen2017person} datasets. For SYSU-MM01 dataset, as we can see from the table, our proposed method works remarkably well, except for the indoor-search scenario on SYSU-MM01, where MRCN \cite{zhang2023mrcn} shows better performance. This method, however, requires to disentangle the features, which is highly time-consuming. For RegDB, our method achieves comparable performance with existing methods and sets a new state of the art. Overall, the experimental results demonstrate that our proposed method provides robust feature representations to the cross-modality discrepancies for image-based VI-reID. We also report the qualitative comparisons along with rank-10 accuracy (\%) in the supplementary material.

\begin{table*}[htbp]
  \centering
  \caption{Quantitative comparison with the state of the art for image-based VI-ReID. We report the rank-1 accuracy (\%) and mAP (\%) on the SYSU-MM01 \cite{wu2017rgb} and RegDB \cite{nguyen2017person} datasets. The best performance obtained by all considered models are marked in bold.}
  \label{sysu}
  \begin{tabular}{l c  cccccccc}
    \toprule
    \multirow{2}{*} {Settings}  & \multicolumn{4}{r}{SYSU-MM01} \cite{wu2017rgb}   & \multicolumn{4}{r}{RegDB} \cite{nguyen2017person}    \\
    \cmidrule(r){3-6}
    \cmidrule(r){7-10}
    &  &   \multicolumn{2}{c}{\textit{All-search}} &   \multicolumn{2}{c}{\textit{Indoor-search}}  &    \multicolumn{2}{c}{\textit{Visible to Infrared}} &  \multicolumn{2}{c}{\textit{Infrared to Visible}}     \\
    \cmidrule(r){1-2}
    \cmidrule(r){3-4}
    \cmidrule(r){5-6}
    \cmidrule(r){7-8}
    \cmidrule(r){9-10}
    Method & Venue                   & rank-1 &  mAP                 &   rank-1 & mAP      &  rank-1 & mAP       & rank-1  & mAP  \\
    \Xhline{0.4pt}
    %                                             & All-search         & Indoor-search     & Visible to Infrared  & Infrared to Visible
    Zero-Pad \cite{wu2017rgb}        & ICCV'17     & 14.80  & 15.95     &20.58 &26.92       &17.75 & 18.90         &16.63 &17.82 \\
    HCML \cite{ye2018hierarchical}   & AAAI'18     & 14.32 & 16.16       & 24.52 &30.08      & 24.44 &20.08      & 21.70  & 22.24 \\
    AlignGAN \cite{wang2019rgb}      & ICCV'19     & 42.40 & 40.70       &45.90 &54.30       & 57.90 & 53.60      & 56.30 & 53.40 \\
    Xmodal \cite{li2020infrared}     & AAAI'20     & 49.92 & 50.73      & - &   -           & 62.21  & 60.18       &  68.06 & 61.80 \\
    DDAG \cite{ye2020dynamic}        & ECCV'20     & 54.75 & 53.02      & 61.02  & 67.98    & 69.34 & 63.46         & 68.06 & 61.80 \\
    LbA \cite{park2021learning}      & ICCV'21     &  55.41 & 54.14      & 58.46 & 66.33     & 74.17  &67.64         & 72.43 & 65.46 \\
    NFS \cite{chen2021neural}        & CVPR'21    & 56.91   & 55.45     &62.79  & 69.79    & 80.54  & 72.10         & 77.95 &69.79  \\
    MC-AWL \cite{ling2021multi}      & IJCAI'21   & 64.82  &  60.81      & 68.05 &  51.48      &93.83 &87.55         & 91.55 &85.25 \\
    MID \cite{huang2022modality}     & AAAI'22    & 60.27   & 59.40     & 64.86 & 70.12       &87.45   & 84.85         & 84.29 & 81.41 \\
    SPOT   \cite{chen2022structure}  &TIP'22      & 65.34   & 62.25       &69.42  & 74.63       & 80.35 & 72.46          & 79.37 &72.26  \\
    DART \cite{yang2022learning}     & CVPR'22    &68.72 &66.29          &72.52 & 78.17      &83.60 &75.67               &81.97 & 73.78 \\
    FMCNet \cite{zhang2022fmcnet}    & CVPR'22    & 66.30   &  62.50      & 68.20 &  74.10       & 89.10  & 84.40         & 88.40 & 83.90 \\
    MRCN \cite{zhang2023mrcn}        & AAAI'23    & 70.80   & 67.30      & 76.40 &  \textbf{80.00}      & 95.10  & 89.20         & 92.60 & 86.50 \\
   \midrule
   Ours  &  - & \textbf{72.56} &\textbf{68.92} & \textbf{76.57}& 78.20 &\textbf{97.53} &\textbf{ 92.97} & \textbf{94.82} & \textbf{89.84}  \\
    \bottomrule
  \end{tabular}
\end{table*}

\begin{table*}[htbp]
  \centering
  \caption{Comparison with the state-of-the-arts for video-based HITSZ-VCM \cite{lin2022learning} dataset. Rank-k accuracy (\%) and mAP (\%) are reported. The best results are marked in bold.}
  \label{HITSZ}
  \begin{tabular}{l c cc c ccccc}
    \toprule
      \multirow{2}{*} {Method}  & \multirow{2}{*} {Venue}    & \multicolumn{4}{c}{\textit{Infrared to Visible}} & \multicolumn{4}{c}{\textit{Visible to Infrared}}   \\
    \cmidrule(r){3-6}
    \cmidrule(r){7-10}
     &    &   rank-1 &  rank-10  &  rank-20 & mAP            & rank-1 &  rank-10 &  rank-20 & mAP  \\
    \Xhline{0.4pt}
    %                               & rank-1  & rank-10  &  rank-20 & mAP    & rank-1  & rank-10 &  rank-20 & mAP
     DDAG \cite{ye2020dynamic}       & ECCV'20 & 54.62  &  76.05 & 81.50 & 39.26     & 59.03  & 79.53 &84.04 & 41.50  \\
     LbA \cite{park2021learning}     & ICCV'21 & 46.38 & 72.23 &79.41 & 30.69        & 49.30  & 75.90 &82.21 & 32.38 \\
     MPANet \cite{wu2021discover}    & CVPR'21 & 46.51  &  70.51 &77.77 &  35.26     & 50.32   & 73.56 & 79.66  & 37.80  \\
     VSD \cite{tian2021farewell}     & CVPR'21 & 54.53  & 76.28 &82.01 &  41.18      & 57.52   & 79.38 &83.61 & 43.45 \\
     CAJL \cite{ye2021channel}       & ICCV'21 & 56.59  &  79.52 & 84.05 & 41.49      & 60.13   & 79.86 & 84.53 & 42.81  \\
    %\Xhline{0.4pt}
     MITML \cite{lin2022learning}    & CVPR'22 & 63.74  & 81.72 &86.28 & 45.31       & 64.54   & 82.98 & 87.10 &  47.69 \\
     IBAN  \cite{li2023intermediary} & TCSVT'23 & 65.03  & 82.98 &87.19 & 48.77     & 69.58   & 85.43 &88.78 & 50.96  \\
   \midrule
     Ours  & -     & \textbf{68.82} & \textbf{85.69}&\textbf{89.31} & \textbf{52.64}     & \textbf{71.62} & \textbf{86.87}& \textbf{89.94} & \textbf{53.16}   \\
   \bottomrule
  \end{tabular}
\end{table*}

\paragraph{Results on HITSZ-VCM.} We compare our model with existing VI-ReID methods including DDAG \cite{ye2020dynamic}, LbA \cite{park2021learning}, MPANet \cite{wu2021discover}, VSD \cite{tian2021farewell}, CAJL \cite{ye2021channel}, MITML \cite{lin2022learning} and IBAN  \cite{li2023intermediary}. Table \ref{HITSZ} presents the rank-k accuracy (\%) and mAP (\%) for the video-based VI-ReID HITSZ-VCM dataset. Note that, only MITML  and IBAN are developed for video-based VI-ReID dataset, while other methods are primarily designed for image-based datasets. From the table, we can see that our model outperforms all the existing methods. Specifically, for the infrared to visible retrieval mode, our model acquires 3.79\% and 3.87\% improvements on rank-1 and mAP respectively than the second best method IBAN. As for visible to infrared mode, we also get a better performance, indicating the effectiveness of our proposed method.

\subsection{Ablation Study}

\begin{table*}[htbp]
  \centering
  \caption{Comparison with variants of our model on the SYSU-MM01 dataset under all search mode.}
  \label{Ablat}
  \begin{tabular}{l c cc c c cccc}
    \toprule
      \multirow{3}{*} {Exp}  & \multicolumn{4}{c}{Alignment with $A$} & \multirow{3}{*} {rank-1} & \multirow{3}{*} {rank-5} & \multirow{3}{*} {rank-10}  & \multirow{3}{*} {rank-20} &\multirow{3}{*}  {mAP} \\
    \cmidrule(r){2-5}
      & \multicolumn{3}{c}{PHO} & \multirow{2}{*} {SGD}  & &  \\
    \cmidrule(r){2-4}
     &      SAS &  AAL  &  CCL &  &         &  &   &   \\
    \Xhline{0.4pt}
     1(base)       &   &   &  &                      & 66.85  & 88.25 &94.31  & 97.84 & 62.36 \\
     2          &   &   &  &  \ding{51}              & 67.21  & 88.56 &94.42 & 97.92 & 62.41 \\
        %\midrule
     3        & \ding{51}  &   & \ding{51} &         & 72.05   & 90.85 & 95.67  & 98.32 & 68.25 \\
     4      &   &  \ding{51}  &  \ding{51} &         & 72.56   & 91.61 &96.19 & 98.65 & 68.92\\
   \bottomrule
  \end{tabular}
\end{table*}

In this section, we conduct ablation study to show the contribution of each strategy proposed in our PHO method. All the experiments
are tested on SYSU-MM01 dataset with all search mode. We first introduce the baseline (Exp 1), which is similar to \cite{jambigi2021mmd} with identity loss, hetero-center triplet loss and maximum mean discrepancy loss. It also means that the issue of cross-modality alignment is not considered by baseline. As shown in Section \ref{Hier}, the cross-modality alignment can be solved by transformation matrix $A$. In Exp 2, we regard the transformation matrix $A$ as the learnable parameters and train the whole network to obtain the optimal values. In Exp 3 and 4, we use parameter hierarchical optimization method to directly obtain the optimal values of transformation matrix $A$ instead of training the whole network. As shown in Table \ref{Ablat}, parameter hierarchical optimization gets a better results, which means explicitly optimize part of parameters via optimization principles instead of training can yield a better manner of parameters training. The above ablation study demonstrate the effectiveness of our proposed method.
In addition, from the Exp 3 and 4, we can see that auto-weighted learning is a little better than the self-adaptive alignment strategy. This is because different features have different importance. Without the cross-modality consistent learning loss, self-adaptive alignment strategy and auto-weighted learning can not work, while using all the losses and strategies gives the best performance, suggesting that they are complementary to each other.

\subsection{Visualization of feature distribution.}

\begin{figure*}[htbp]
\begin{center}
%\centerline{\includegraphics[height=4cm,width=7cm]{DNA.pdf}}
\centerline{\includegraphics[width=12cm]{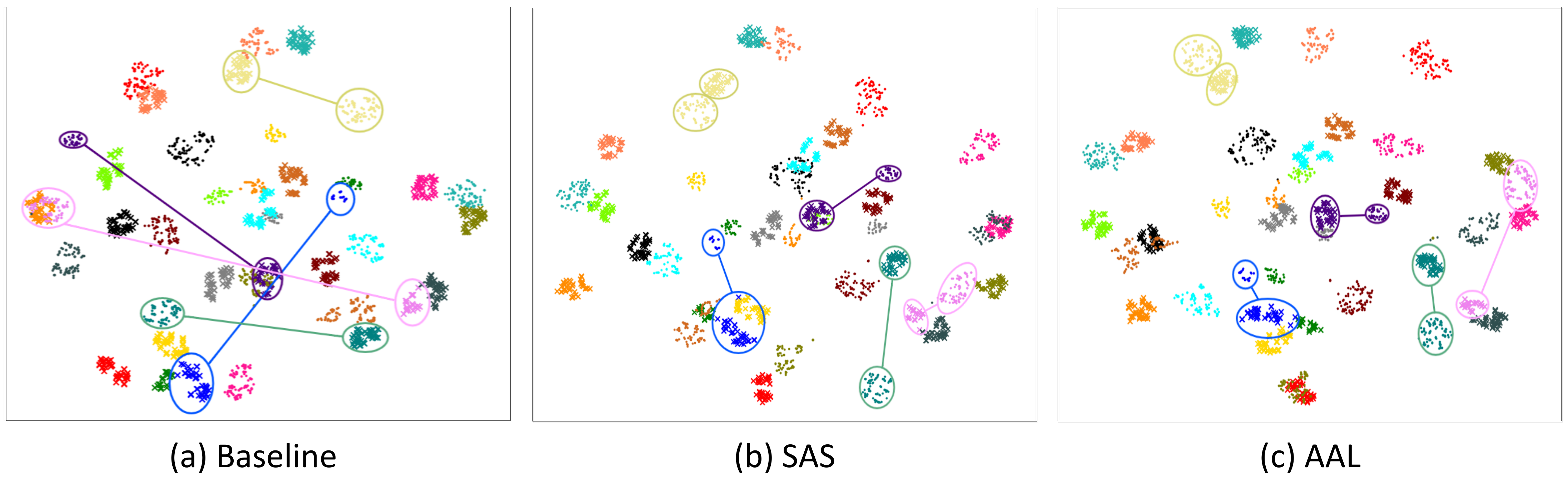}}
\caption{Visualization of learned features with t-SNE method.}
\label{tsne}
\end{center}
\end{figure*}

We use the t-SNE \cite{van2008visualizing} to visualize feature distribution of PHO person representations in 2D space on SYSU-MM01 dataset.
The same color indicates the same person, and the different shapes represent different modalities. As shown in Fig. \ref{tsne}, the baseline can not well separate the different persons, while the same identity with different modalities departs away from each other. However, our method can better aggregate discriminative feature representations of cross-modality, and make the same person with different modalities close to each other rather than different persons.

\section{Conclusions}

In this paper, we propose a novel parameter optimizing manner PHO method. Specifically, to address the cross-modality alignment, we propose a self-adaptive alignment strategy, which can automatically align the visible and infrared images. We also design the auto-weighted alignment learning to automatically weight features according to their importance. It should be pointed out that in the process of SAS and AAL, all the parameters are directly optimized with optimization rules rather than training the whole network, which can narrow the search space of parameters and make the remaining parameters easier to be trained. Furthermore, to extract discriminative person representations, the cross-modality consistent learning loss is introduced based on translation consistency. Experimental results on image- and video-based VI-reID datasets show the effectiveness of our proposed method.

\section{Appendices}

Denote that\\
\vspace{2mm}
\quad \quad \quad \quad $A=\left(\begin{array}{cccc}A_{11} & A_{12} & \cdots & A_{1n}\\
A_{21} & A_{22} & \cdots & A_{2n}\\
\vdots & \vdots & \vdots & \vdots\\
A_{n1} & A_{n2} & \cdots & A_{nn}
\end{array}\right)$, %\\
\vspace{2mm}
 $W(n)=\left(\begin{array}{cccc}w_{1}(n) & 0 & \cdots & 0\\
0 & w_{2}(n) &  \ddots &  \vdots\\
\vdots & \ddots &  \ddots & 0\\
0 &  \cdots & 0 & w_{n}(n)
\end{array}\right)$,\\
$ \bar{X}= (\bar{X}_{1},\bar{X}_{2},...,\bar{X}_{n})^T$ and $ \bar{Y}= (\bar{Y}_{1},\bar{Y}_{2},...,\bar{Y}_{n})^T$.
Then, the problem $P_{2}(W(n),A)$ can be rewritten as
\begin{flalign*}
& P_{2}\big(W(n),A\big)= \left\Vert W(n) \left( \frac{1}{M} \sum_{i=1}^{M} AX_{i}- \frac{1}{N}\sum_{j=1}^{N} Y_{j} \right) \right\Vert_2^2 + \alpha \left\Vert A\right\Vert_F^2 &
\end{flalign*}
\begin{flalign*}
& =\left\| \left(\begin{array}{cccc}w_{1}(n) & 0 & \cdots & 0\\
0 & w_{2}(n) &  \ddots &  \vdots\\
\vdots & \ddots &  \ddots & 0\\
0 &  \cdots & 0 & w_{n}(n)
\end{array}\right) \right.
\times
%\end{flalign*}
%\begin{flalign*}
 \left[\left(\begin{array}{cccc}A_{11} & A_{12} & \cdots & A_{1n}\\
A_{21} & A_{22} & \cdots & A_{2n}\\
\vdots & \vdots & \vdots & \vdots\\
A_{n1} & A_{n2} & \cdots & A_{nn}
\end{array}\right) \times
\left(\begin{array}{c}
\bar{X}_{1}\\
\bar{X}_{2} \\
\vdots \\
\bar{X}_{n}
\end{array}\right)\right. &
\end{flalign*}
\begin{flalign*}
& \left.\left.
\ \ \ \ -\left(\begin{array}{c}
\bar{Y}_{1}\\
\bar{Y}_{2} \\
\vdots \\
\bar{Y}_{n}
\end{array}\right)\right]\right\|_{2}^{2} +
%\alpha \sum_{i=1}^{n}\sum_{j=1}^{n}A^2_{ij}
%
 \alpha \left\| \left( \begin{array}{cccc}A_{11} & A_{12} & \cdots & A_{1n}\\
A_{21} & A_{22} & \cdots & A_{2n}\\
\vdots & \vdots & \vdots & \vdots\\
A_{n1} & A_{n2} & \cdots & A_{nn}
\end{array}\right) \right\|_F^2
&
\end{flalign*}
\begin{flalign*}
& =\sum_{i=1}^{n}w_{i}^{2}(n)\left(\sum_{j=1}^{n}A_{ij}\bar{X}_{j}-\bar{Y}_{i}\right)^{2} + \alpha \sum_{i=1}^{n}\sum_{j=1}^{n}A^2_{ij} &
\end{flalign*}

\subsection{Proof of Theorem 3}

\textit{proof.}
When $W(n)=\hat{W}(n)$ is fixed, if $A^\star= (A_{ij}^\star )_{n\times n}$ is to minimize $P_{2}\big(\hat{W}(n),A\big)$, its gradient in both sets of variables must vanish. Hence,
\begin{equation}
\label{A1}
\left\{\begin{array}{c}
\frac{\partial P_{2}\big(\hat{W}(n),A^\star\big)}{\partial A^\star_{11}}=2\hat{w}_{1}^{2}(n)\left(\sum_{j=1}^{n}A^\star_{1j}\bar{X}_{j}-\bar{Y}_{1}\right)\bar{X}_{1} + 2 \alpha A^\star_{11} \\
\frac{\partial P_{2}\big(\hat{W}(n),A^\star\big)}{\partial A^\star_{12}}=2\hat{w}_{1}^{2}(n)\left(\sum_{j=1}^{n}A^\star_{1j}\bar{X}_{j}-\bar{Y}_{1}\right)\bar{X}_{2} + 2 \alpha A^\star_{12}\\
\vdots \\
\frac{\partial P_{2}\big(\hat{W}(n),A^\star\big)}{\partial A^\star_{1n}}=2\hat{w}_{1}^{2}(n)\left(\sum_{j=1}^{n}A^\star_{1j}\bar{X}_{j}-\bar{Y}_{1}\right)\bar{X}_{n} + 2 \alpha A^\star_{1n} 
\end{array}
\right.
\end{equation}
\begin{equation}
\label{A2}
\left\{\begin{array}{c}
\frac{\partial P_{2}\big(\hat{W}(n),A^\star\big)}{\partial A^\star_{21}}=2\hat{w}_{2}^{2}(n)\left(\sum_{j=1}^{n}A^\star_{2j}\bar{X}_{j}-\bar{Y}_{2}\right)\bar{X}_{1} + 2 \alpha A^\star_{21} \\
\frac{\partial P_{2}\big(\hat{W}(n),A^\star\big)}{\partial A^\star_{22}}=2\hat{w}_{2}^{2}(n)\left(\sum_{j=1}^{n}A^\star_{2j}\bar{X}_{j}-\bar{Y}_{2}\right)\bar{X}_{2} + 2 \alpha A^\star_{22} \\
\vdots \\
\frac{\partial P_{2}\big(\hat{W}(n),A^\star\big)}{\partial A^\star_{2n}}=2\hat{w}_{2}^{2}(n)\left(\sum_{j=1}^{n}A^\star_{2j}\bar{X}_{j}-\bar{Y}_{2}\right)\bar{X}_{n} + 2 \alpha A^\star_{2n} 
\end{array}
\right.
\end{equation}

\begin{equation}
\vdots    \nonumber
\end{equation}

\begin{equation}
%\label{An}
\left\{\begin{array}{c}
\frac{\partial P_{2}\big(\hat{W}(n),A^\star\big)}{\partial A^\star_{n1}}=2\hat{w}_{n}^{2}(n)\left(\sum_{j=1}^{n}A^\star_{nj}\bar{X}_{j}-\bar{Y}_{n}\right)\bar{X}_{1} + 2 \alpha A^\star_{n1} \\
\frac{\partial P_{2}\big(\hat{W}(n),A^\star\big)}{\partial A^\star_{n2}}=2\hat{w}_{n}^{2}(n)\left(\sum_{j=1}^{n}A^\star_{nj}\bar{X}_{j}-\bar{Y}_{n}\right)\bar{X}_{2} + 2 \alpha A^\star_{n2}\\
\vdots \\
\frac{\partial P_{2}\big(\hat{W}(n),A^\star\big)}{\partial A^\star_{nn}}=2\hat{w}_{n}^{2}(n)\left(\sum_{j=1}^{n}A^\star_{nj}\bar{X}_{j}-\bar{Y}_{n}\right)\bar{X}_{n} + 2 \alpha A^\star_{nn}
\end{array}
\right.
\end{equation}

Let $T_{1}=-\frac{1}{\alpha}\hat{w}_{1}^{2}(n)\left(\sum_{j=1}^{n}A^\star_{1j}\bar{X}_{j}-\bar{Y}_{1}\right)$.
%and $T_{n}=-\frac{1}{\alpha} \hat{w}_{n}^{2}(n)\left(\sum_{j=1}^{n}A^\star_{nj}\bar{X}_{j}-\bar{Y}_{1}\right)$.
It follows from equation (\ref{A1}) %and (\ref{An})
\begin{equation}
\label{T1}
\left\{\begin{array}{c}
A^\star_{11}=T_{1}\bar{X}_{1}\\
A^\star_{12}=T_{1}\bar{X}_{2}\\
\vdots \\
A^\star_{1n}=T_{1}\bar{X}_{n}
\end{array}
\right.
\end{equation}

Substituting (\ref{T1}) into (\ref{A1}), we get
\begin{equation}
\label{qqqq}
\left\{\begin{array}{c}
 \hat{w}_{1}^{2}(n)\left(\sum_{j=1}^{n}T_{1}\bar{X}_{j}\bar{X}_{j}-\bar{Y}_{1}\right)\bar{X}_{1} +  \alpha T_{1}\bar{X}_{1} \\
 \hat{w}_{1}^{2}(n)\left(\sum_{j=1}^{n}T_{1}\bar{X}_{j}\bar{X}_{j}-\bar{Y}_{1}\right)\bar{X}_{2} + \alpha T_{1}\bar{X}_{2} \\
\vdots \\
 \hat{w}_{1}^{2}(n)\left(\sum_{j=1}^{n}T_{1}\bar{X}_{j}\bar{X}_{j}-\bar{Y}_{1}\right)\bar{X}_{n} + \alpha T_{1}\bar{X}_{n} 
\end{array}
\right.
\end{equation}

From (\ref{qqqq}), we obtain
\begin{equation}
T_{1} = \frac{\hat{w}_{1}^{2}(n) \bar{Y}_{1}}{  \hat{w}_{1}^{2}(n) \sum_{j=1}^{n} \bar{X}_{j}^2 + \alpha }
\end{equation}

Therefore
\begin{equation}
\left(\begin{array}{c}
A^\star_{11}\\
A^\star_{12}\\
\vdots \\
A^\star_{1n}
\end{array} \right)
=\frac{\hat{w}_{1}^{2}(n) \bar{Y}_{1}}{  \hat{w}_{1}^{2}(n) \sum_{j=1}^{n} \bar{X}_{j}^2 + \alpha }
\left(\begin{array}{c}
\bar{X}_{1}\\
\bar{X}_{2}\\
\vdots \\
\bar{X}_{n}
\end{array}\right)
\end{equation}

Similarly, we have
\begin{equation}
A_{ij}^\star= \frac{\hat{w}_{i}^{2}(n) \bar{Y}_{i}\bar{X}_{j}}{  \hat{w}_{i}^{2}(n) \sum_{p=1}^{n} \bar{X}_{p}^2 + \alpha } \ \ for \ 1\leq i,j\leq n.
\end{equation}

\subsection{Proof of Theorem 2}

\textit{proof.}
Let $w_{i}(n)=\hat{w}_{i}(n)=1, i = 1,2,...,n$. The proof is similar to Theorem 3.

\subsection{Proof of Theorem 4}

\textit{proof.}
When $A =\hat{A}$ is fixed, we can rewrite problem $P_{2}\big(W(n),\hat{A}\big)$ as

\begin{flalign*}
P_{2}\big(W(n),\hat{A}\big) & = \sum_{i=1}^{n}w_{i}^{2}(n)\left(\sum_{j=1}^{n}\hat{A}_{ij}\bar{X}_{j}-\bar{Y}_{i}\right)^{2} + \alpha \sum_{i=1}^{n}\sum_{j=1}^{n}\hat{A}^2_{ij}  & \\
& = \sum_{i=1}^{n}w_{i}^{2}(n)E_{i} + \alpha \sum_{i=1}^{n}\sum_{j=1}^{n}\hat{A}^2_{ij} &
\end{flalign*}
where $E_{1},E_{2},...,E_{n}$ are $n$ constants when $A =\hat{A}$ is fixed.

If $E_{i}=0$, it means that the $i$-th dimension has no contribution to the objective function. For convenience, we set $w^{*}_{i}(n)=0$ to the $i$-$th$ dimension where $E_{i}=0$.

If $E_{i}\neq 0$, we can use the Lagrange multiplier method to solve the problem $P_{2}(W(n),\hat{A})$ with constraint condition $\sum_{i=1}^{n}w_{i}(n)=1$. Let $\tau$ be the Lagrange multiplier and
\begin{equation}
\Phi(W(n),\tau)=\sum_{i=1}^{n}w_{i}^{2}(n)E_{i} + \alpha \sum_{i=1}^{n}\sum_{j=1}^{n}\hat{A}^2_{ij} + \tau \left(\sum_{i=1}^{n}w_{i}(n)-1\right)
\end{equation}

If $ \big(W^\star(n),\tau^\star \big)$ is to minimize $\Phi(W(n),\tau)$, the gradients of all variables must vanish. Hence
\begin{equation}
\label{LagrW}
\frac{\partial \Phi \big(W^\star(n),\tau^\star \big)}{\partial w_{i}^\star(n) }=2 w_{i}^\star(n) E_{i}+ \tau^\star =0
\end{equation}
\begin{equation}
\label{LagrLam}
\frac{\partial \Phi \big(W^\star(n),\tau^\star \big)}{\partial \tau^\star  }=\sum_{i=1}^{n}w^\star_{i}(n)-1=0
\end{equation}

From (\ref{LagrW}), we have
\begin{equation}
\label{slov}
w^\star_{i}(n)= \frac{-\tau^\star}{2E_{i}} \ \  \ \ \textrm{\emph{for all}}\ 1\leq i\leq n
\end{equation}

Substituting (\ref{slov}) into (\ref{LagrLam}), we obtain
\begin{equation}
\label{tao}
\sum_{t=1}^{n} \frac{-\tau^\star}{2E_{t}} =1
\end{equation}

We derive form (\ref{tao})
\begin{equation}
\label{tao11}
\tau^\star =\frac{-2}{\sum_{t=1}^{n} \frac{1}{E_{t}}}
\end{equation}

Hence, it follows from (\ref{slov}) and (\ref{tao11}) that
\begin{equation}
  w_{i}^\star(n)= \frac{1}{\mathop{ \sum_{t=1}^{n}}\limits_{E_{t}\neq 0} \frac{E_{i}}{E_{t}}}
\end{equation}

 \section{Qualitative comparisons}

 We provide the comparison results with the state-of-the-art methods on SYSU-MM01 and RegDB datasets, respectively.

\begin{table*}[htbp]
  \centering
  \caption{Quantitative results of rank-k accuracy (\%) on SYSU-MM01 and RegDB datasets. }
  %\label{sysu}
  \begin{tabular}{l  ccccccc}
    \toprule
     \multicolumn{4}{c}{SYSU-MM01} & \multicolumn{4}{c}{RegDB}  \\
    \cmidrule(r){1-4}
    \cmidrule(r){5-8}
    \multicolumn{2}{c}{\textit{All-search}} &  \multicolumn{2}{c}{\textit{Indoor-search} } &   \multicolumn{2}{c}{\textit{Visible to Infrared}} &  \multicolumn{2}{c}{\textit{Infrared to Visible} }    \\

    \cmidrule(r){1-2}
    \cmidrule(r){3-4}
    \cmidrule(r){5-6}
    \cmidrule(r){7-8}
    rank-10 &   rank-20 &  rank-10 &rank-20 & rank-10 &   rank-20 &  rank-10 &rank-20 \\
    \Xhline{0.4pt}
    %                                             & All-search         & Indoor-search     & Visible to Infrared  & Infrared to Visible

      96.19  &  98.65 &  98.62  & 99.24 &  98.91  & 99.53 &98.13  & 99.64 \\
    \bottomrule
  \end{tabular}
\end{table*}

%------------------------------------------------------------------------

{\small
\bibliographystyle{ieee}
\bibliography{egbib}
}

%\begin{thebibliography}{00}
%\bibitem{b1} G. Eason, B. Noble, and I. N. Sneddon, ``On certain integrals of Lipschitz-Hankel type involving products of Bessel functions,'' Phil. Trans. Roy. Soc. London, vol. A247, pp. 529--551, April 1955.
%\bibitem{b2} J. Clerk Maxwell, A Treatise on Electricity and Magnetism, 3rd ed., vol. 2. Oxford: Clarendon, 1892, pp.68--73.
%\bibitem{b3} I. S. Jacobs and C. P. Bean, ``Fine particles, thin films and exchange anisotropy,'' in Magnetism, vol. III, G. T. Rado and H. Suhl, Eds. New York: Academic, 1963, pp. 271--350.
%\bibitem{b4} K. Elissa, ``Title of paper if known,'' unpublished.
%\bibitem{b5} R. Nicole, ``Title of paper with only first word capitalized,'' J. Name Stand. Abbrev., in press.
%\bibitem{b6} Y. Yorozu, M. Hirano, K. Oka, and Y. Tagawa, ``Electron spectroscopy studies on magneto-optical media and plastic substrate interface,'' IEEE Transl. J. Magn. Japan, vol. 2, pp. 740--741, August 1987 [Digests 9th Annual Conf. Magnetics Japan, p. 301, 1982].
%\bibitem{b7} M. Young, The Technical Writer's Handbook. Mill Valley, CA: University Science, 1989.
%\end{thebibliography}

\end{document}